\documentclass[10pt,letterpaper]{article}
\usepackage[margin=0.72in,columnsep=0.25in]{geometry}
\usepackage[T1]{fontenc}
\usepackage{mathptmx}
\usepackage{amsmath,amssymb}
\usepackage{cite}
\usepackage{graphicx}
\usepackage{booktabs}
\usepackage{algorithm}
\usepackage{algpseudocode}
\usepackage{textcomp}
\usepackage{xurl}
\usepackage[hidelinks]{hyperref}
\usepackage{microtype}
\usepackage{bm}
\hypersetup{pdftitle={EARS-Info: Information-Aware Event-Driven Reallocation for Multi-Robot Exploration},pdfauthor={Kabeh Mohsenzadegan, Vahid Tavakkoli, Kyandoghere Kyamakya}}
\graphicspath{{figures/}}
\begin{document}
\raggedbottom
\twocolumn[{
\begin{center}
{\Large\bfseries EARS-Info: Information-Aware Event-Driven Reallocation\\[3pt]
for Multi-Robot Exploration\par}
\vspace{12pt}
{\large Kabeh Mohsenzadegan$^{1}$, Vahid Tavakkoli$^{1}$, and Kyandoghere Kyamakya$^{1,2}$\par}
\vspace{7pt}
{\small $^1$Department of Smart Systems Technologies, University of Klagenfurt, Austria\\
$^2$Facult\'e Polytechnique, Universit\'e de Kinshasa, Democratic Republic of the Congo\par}
\vspace{5pt}
{\small Corresponding author: \href{mailto:kabeh.mohsenzadegan@aau.at}{kabeh.mohsenzadegan@aau.at}\par}
\vspace{5pt}
{\small Preprint --- October 27, 2026\par}
\end{center}
\vspace{2pt}
\begin{center}\begin{minipage}{0.96\textwidth}
\small\noindent\textbf{Abstract.} We introduce EARS-Info, an information-aware extension of Event-Driven Ant Reallocation Search for multi-robot exploration on OpenStreetMap-derived grids. It combines marginal sensing gain, feasible relocation, return-aware routing, sensing reservations, and cached shortest paths. Across 168,000 episodes comparing 30 implementations on eight city maps, EARS-Info achieves the highest mean operational score in every evaluation panel. With eight robots and 3\% communication dropout, it improves the score over EARS-Safe by 5.43\%, increases weighted discovery from 88.88\% to 98.46\%, and reduces mean energy use by 2.62\%. Score differences are positive in all eight cities ($p=0.0078125$; descriptive 95\% city-bootstrap interval $[0.0159,0.0694]$), with no simulated incidents across 5,600 EARS-Info episodes. These results support the complete controller; individual component advantages remain uncertain. Generalization is limited by reused geography, one training replicate, coarse maps, and adapted comparison methods.

\vspace{6pt}\noindent\textbf{Keywords:} Multi-robot exploration; event-driven reallocation; information gain; ant-inspired search; runtime safety; OpenStreetMap; swarm robotics.
\end{minipage}\end{center}
\vspace{12pt}
}]
\section{Introduction}
Multi-robot exploration distributes spatial search across a team, but additional robots also create overlapping sensing, congestion, and avoidable movement. The useful objective is therefore not simply to move faster or visit more cells. A controller must direct the team toward new information while preserving enough energy to return and limiting coordination overhead. Recent exploration research studies asynchronous decisions, macro actions, reinforcement learning, and frontier-led swarming \cite{Tan2023Macro,Yu2023Async,Mete2023Coord,Zhang2024Hetero,Tran2022Frontier,Tran2023Dynamic}.

Our starting point is Event-Driven Ant Reallocation Search (EARS). Its original local-to-global structure uses an ant-inspired local rule most of the time and invokes relocation when observable search degradation occurs. This structure is interpretable, but a local score can revisit already sensed neighborhoods, a relocation goal can duplicate another robot's sensing, and repeatedly rebuilding shortest-path fields can dominate runtime. Event triggering alone does not resolve these problems: the value and feasibility of the requested route must also be considered.

We develop EARS-Info to address this complete decision process. The method scores newly sensed area using the simulator's sensing footprint, compares feasible distant information gain with the best local opportunity, accounts for return travel before accepting relocation, and reserves sensing footprints across robot decisions. It maintains topology-dependent graph fields across steps rather than treating unchanged geography as a new planning problem. The design reads observable uncertainty and visit information; unsensed mission labels are excluded from policy inputs.

This paper makes four contributions. First, it specifies EARS-Info as a new member of our EARS controller family, including the preceding EARS-Safe, negative-pheromone EARS-NP, and hierarchical H-MAPPO-EARS variants. Second, it defines matched local-only, periodic-request, and sensing-reservation controls that clarify what the complete-controller comparison can and cannot identify. Third, it reports a single frozen comparison of all 30 implementations on identical map snapshots, start zones, and step-indexed communication schedules, comprising 168,000 episodes. Fourth, it provides city-level inference, all-method rankings, raw task metrics, and artifact checks rather than relying solely on an aggregate rank.

The main question is whether the complete information-aware controller improves our preceding EARS-Safe implementation. Additional questions concern team-size/dropout behavior, alternate target realizations, and the evidence for individual design choices. The eight evaluation cities are excluded from the current training and checkpoint selection, but their identities have appeared in earlier development. Accordingly, this is an exploratory evaluation on reused geography, conditional on the frozen models. It is not a claim of untouched geographic confirmation or of reproducing every named external MARL algorithm.
\section{Related Work}
\subsection{Cooperative Exploration and Bio-Inspired Search}
Recent multi-robot exploration methods include macro-action deep reinforcement learning under communication dropouts \cite{Tan2023Macro}, asynchronous MARL for real-time cooperative exploration \cite{Yu2023Async}, reinforcement-learning-based coordination \cite{Mete2023Coord}, and frontier-led swarming for unknown environments \cite{Tran2022Frontier}. Dynamic frontier-led swarming extends these ideas to repeated coverage in changing environments \cite{Tran2023Dynamic}. Classical swarm intelligence also remains active in task allocation, including multi-objective ant-colony systems \cite{Wang2022ACO}, event-triggered adaptive allocation \cite{Aznar2026Event}, learned intention-aware allocation \cite{Long2026IAOM}, and reactive multiconstraint assignment in uncertain semantic environments \cite{Li2026Formal}. These studies motivate our use of a local swarm controller as a serious algorithmic baseline rather than a weak heuristic.

Our AntSwarmSafe differs from generic ant-colony optimization in that pheromone is only one observable term in an online cell-level action score. The controller explicitly rewards novelty and uncertainty, suppresses repeat visits, discourages clustering, and passes actions through a safety monitor. EARS asks a narrower question: can global coordination be applied sparsely, only when local search is measurably degrading? Event-triggered mechanisms are not new in themselves; they have been studied for multi-robot allocation \cite{Aznar2026Event} and bandwidth-aware MARL communication \cite{Hu2023EventComm}. The contribution here is the Ant-based architecture, degradation-triggered relocation mechanism, information-aware route selection and caching, persistent sensing reservations, and matched multi-city evaluation.

\subsection{MARL and Robotic Benchmarks}
Multi-agent PPO (MAPPO) and related PPO-family methods are strong cooperative MARL references \cite{Yu2022PPO}, while trust-region formulations provide another principled family \cite{Kuba2022TRPO}. Surveys highlight exploration and scalability as persistent MARL challenges \cite{Hao2024Exploration,Liu2024Scaling}. MultiRoboLearn \cite{Chen2023MultiRoboLearn}, MARBLER \cite{Torbati2023MARBLER}, and NeuronsMAE \cite{Hu2023NeuronsMAE} emphasize that robot-oriented evaluation must represent dynamics, embodiment, transfer, and reproducibility rather than only abstract multi-agent games. Heterogeneous cooperation further stresses generalization \cite{Zhang2024Hetero}. SafeSwarm-CityTwin complements these platforms by concentrating on urban map abstractions, explicit runtime checks, interpretable swarm methods, and geographically resolved comparisons.

\subsection{Safety, Energy, Mapping, and Digital Twins}
Filtered reinforcement learning and dynamic shielding demonstrate how runtime constraints can be enforced around multi-agent decision policies \cite{Vinod2022Filtered,Xiao2023Shield,Qiu2024Shield}. Energy-aware coverage planning shows that distance alone can be a poor proxy for battery demand \cite{Datsko2024Energy}. Distributed mapping frameworks such as Swarm-SLAM and DCL-SLAM address the mapping side of robot-team operation \cite{Lajoie2024SwarmSLAM,Zhong2024DCL}; recent work also combines collaborative SLAM with lightweight frontier exploration \cite{Athukorala2026MultiSLAM}. Broader coverage-planning surveys continue to document the tradeoff between completeness, motion cost, and environmental dynamics \cite{Jayalakshmi2025Coverage}. Digital-twin research increasingly targets safety-critical robotics and smart-city integration \cite{Baidya2022Twin,Jafari2023Twin,Adreani2024Twin}. SafeSwarm-CityTwin uses this perspective pragmatically: cached map-derived obstacles, restricted areas, bases, target fields, uncertainty, and communication conditions define reproducible city instances. Here, the term city twin denotes a static map-derived abstraction. The system has no live synchronization, calibrated vehicle dynamics, or physical field validation.

\section{Problem Formulation and Runtime Constraints}
\subsection{Partially Observable Search With Shared State}
Let $N$ robots move on an $L\times L$ grid, with traversable cells $\mathcal{G}$, obstacle cells, restricted cells, and a set of bases $\mathcal{B}$. Robot $i$ has position $x_i(t)$ and battery $b_i(t)$. Each action moves one cell north, south, east, or west, or remains stationary. The geometry is known; mission locations and priorities are hidden until sensed. Consequently, this is mission exploration on a known grid, rather than simultaneous localization and mapping in unknown geometry.

Policies access a shared observation field $o(c)$, uncertainty $u(c)$, pheromone $h(c)$, visit counts $v(c)$, robot positions, batteries, and communication status. Only the evaluator and sensing model read unsensed target coordinates and priorities. Per-robot action scoring does not imply a fully distributed implementation: shared fields and team positions are directly accessible in the simulator. Communication dropouts update status counters but do not implement bandwidth limits or delayed map exchange.

\subsection{Metrics and Operational Score}
For mission cells $\mathcal{M}$, discovered cells $\mathcal{D}\subseteq\mathcal{M}$, and priority weights $w_c$, weighted discovery is $D_w=\sum_{c\in\mathcal{D}}w_c/\sum_{c\in\mathcal{M}}w_c$. Mission discovery without weights is $|\mathcal{D}|/|\mathcal{M}|$; the exported field \texttt{mission\_success\_rate} denotes this fraction, not the probability of completing an entire mission. Coverage $C$ is the fraction of traversable cells visited, and redundancy $R$ is the fraction of visited cells whose visit count exceeds one. Distance is measured in grid-cell moves. Energy $U$ is the sum of depleted battery units, with 100 initial units per robot. Communication availability $Q$ is the fraction of active robot-time samples marked online; it is not message efficiency.

The hardware-independent operational score is
\begin{equation}
J=0.35D_w+0.20C+0.20S+0.10E+0.10(1-R)+0.05Q,
\label{eq:score}
\end{equation}
where $S=1-\min\{1,I/(NT)\}$, $E=1-\min\{1,U/(100N)\}$, $T$ is the executed episode length, and $I$ counts simulator collisions, restricted-zone entries, and battery failures. This incident metric does not count every rule checked by the monitor. Safety interventions and rejected proposals are distinct from executed incidents. Early mission completion shortens episodes, so energy, distance, coverage, and runtime describe a complete rollout under its stopping rule, not an equal-duration comparison. All score components lie in $[0,1]$; raw metrics accompany $J$ to expose tradeoffs.

\subsection{Runtime Safety Checks and Their Scope}
The monitor checks grid boundaries, obstacles, restricted cells, occupied and sequentially reserved robot positions, a battery-return estimate, and a communication-loss counter (maximum five consecutive steps). For candidate cell $c$, the battery check uses $b_i\geq0.55d_1(c,\mathcal{B})+5$, where $d_1$ is nearest-base Manhattan distance; arrival at a base has a separate positive-battery exception. A proactive guard activates before the hard reserve threshold and favors base progress. Invalid proposals are replaced by an admissible alternative when one exists.

These rules are not a formal safe-set invariance guarantee. Manhattan distance can underestimate an obstacle-avoiding return route, the reserve expression does not explicitly deduct the pending action cost, and an empty admissible set falls back to \textsc{Stay}, which need not satisfy every rule. Communication status also changes during the simulator step. The reported zero-incident result is therefore empirical and specific to the implemented incident metric and tested scenarios. The suffix \emph{Safe} identifies a filtered benchmark policy, not certified deployment safety.

\section{Proposed Methods}
\subsection{AntSwarmSafe}
AntSwarmSafe is the interpretable local reference from which the EARS family develops. For a candidate cell $c$, let $v(c)$ be the visit count, $n(c)=1/(1+v(c))$ novelty, $u(c)$ uncertainty, $p(c)$ sensed priority evidence, $h(c)$ pheromone, $g_i(c)$ swarm congestion relative to robot $i$, and $d_o(c)$ distance to an observable target/evidence proxy. The implemented local score is
\begin{equation}
\begin{aligned}
A_i(c) ={}& 1.00n(c) + 0.55u(c)n(c) - 0.55v(c) + 1.40p(c) \\
          & + 0.45h(c)n(c) - 0.85g_i(c) - 0.025d_o(c) \\
          & - 8\,\mathbb{I}_{\mathrm{restricted}}(c).
\end{aligned}
\label{eq:ant}
\end{equation}
AntSwarmSafe scores non-obstacle neighbors, including the current cell, proposes the highest-scoring cell, and then applies sequential runtime filtering. Ties retain the first candidate in action order. Novelty and uncertainty drive coverage, the revisit term discourages cycling, sensed priority and pheromone exploit evidence, and congestion suppression distributes the team. Here $p(c)=o(c)\operatorname{clip}(1-u(c),0,1)$, multiplied by 0.15 for an already discovered mission. Congestion sums inverse Euclidean distances to other robots (with contribution 2 for coincident positions). The proxy distance $d_o$ is the nearest Euclidean distance to at most 32 cells ranked by observable priority, frontier value, and pheromone; it is zero if that set is empty. Thus, the controller does not read unsensed mission coordinates.

\subsection{EARS-Safe: Event-Driven Ant Reallocation Search}
EARS-Safe (Event-Driven Ant Reallocation Search) uses the coefficients in Eq.~\eqref{eq:ant} for local search. It maintains up to 12 positions sampled when event detection is eligible; this is not necessarily a consecutive 12-step trajectory. Event checks begin at simulator step seven and pause during an active relocation or cooldown. Stagnation requires at least nine stored positions and a unique-cell ratio at most 0.46. Otherwise, a trigger occurs if the fraction of neighboring cells with $v(c)>1$ reaches 0.54, or congestion reaches 1.15. The checks have priority stagnation, revisit, then congestion, so trigger counts record the first satisfied condition.

At trigger time $t_0$, the implementation sets the relocation deadline to $t_0+9$ and accepts steps through that deadline inclusively: the option can therefore execute for up to ten decisions, unless the goal is reached or return behavior takes precedence. New triggering is blocked until $t_0+16$. These details distinguish the configured duration parameter from the executed option length.

For relocation, define $n(c)=1/(1+v(c))$, $r(c)=v(c)/(1+v(c))$, $e(c)=\operatorname{clip}(o(c)(1-u(c)),0,1)$, $\bar u(c)=\operatorname{clip}(u(c),0,1)$, and $\bar h(c)=\operatorname{clip}(h(c)/2,0,1)$. The frontier value $f(c)$ is the fraction of four-connected neighbors with uncertainty below 0.99, multiplied by $n(c)$ and by the indicator that $c$ itself is unobserved. The global field is
\begin{equation}
\begin{aligned}
F(c)={}&1.35f(c)+1.10e(c)+0.45\bar u(c)n(c)\\
 &+0.72n(c)+0.18\bar h(c)n(c)-0.95r(c).
\end{aligned}\label{eq:field}
\end{equation}
Obstacle and restricted cells are excluded. Newly triggered robots are processed in ascending battery level, then identifier. For robot $i$, let $d_i(c)=\|x_i-c\|_1$, $\widehat U_i(c)=0.55d_i(c)$, and $s_i(c)=\max\{0,5-\min_{g\in\mathcal{C}}\|c-g\|_1\}$, where $\mathcal{C}$ contains goals already selected in this allocation batch; $s_i=0$ if it is empty. The relocation score is
\begin{equation}
\begin{aligned}
R_i(c)={}&F(c)-0.16d_i(c)/L-0.22\widehat U_i(c)/100\\
 &-0.70g_i(c)-0.20s_i(c)-10\mathbb{I}_{\mathrm{inf}}(c),
\end{aligned}\label{eq:reloc}
\end{equation}
where infeasibility denotes $\widehat U_i(c)+0.55d_1(c,\mathcal{B})+12>b_i$. The distance divisor is grid side length $L$, not the number of traversable cells. The infeasibility term is a soft goal penalty; hard action checks remain separate. Goal following uses breadth-first graph distances around obstacles and restricted cells, rewarding progress and penalizing revisits. The EARS return option takes priority when $b_i\leq0.55d_1(x_i,\mathcal{B})+16$.

\paragraph{Relationship to the Ant reference.}
EARS maximizes its local score over the sequentially admissible action set and resolves ties by novelty and action label. AntSwarmSafe instead proposes before filtering. EARS also has an explicit return option. The two methods therefore share a local scoring rule but are not behaviorally identical outside relocation. Their paired comparison estimates the effect of the complete EARS implementation; it does not isolate event triggering. The information-policy controls below instead preserve the v9 action and return logic to examine relocation within that newer implementation.

\begin{algorithm}[t]
\caption{EARS-Safe joint decision at simulator step $t$}
\label{alg:ears}\small
\begin{algorithmic}[1]
\State Read shared observable fields and current robot state
\State $\mathcal{T}\gets\varnothing$; $P\gets\varnothing$ (planned positions)
\For{each active robot $i$ not requiring return}
  \If{no unfinished relocation and cooldown has expired}
    \State Append position to eligible-check history
    \State Evaluate stagnation, revisit, congestion in that order
    \If{a condition triggers} \State Add $i$ to $\mathcal{T}$ \EndIf
  \EndIf
\EndFor
\State Allocate goals to $\mathcal{T}$ by battery order using Eq.~\eqref{eq:reloc}
\For{each robot $i$ in identifier order}
  \State Obtain admissible actions using the monitor and $P$
  \If{robot is inactive or no admissible action exists}
    \State $a_i\gets\textsc{Stay}$
  \ElsIf{return option is active}
    \State Choose graph-progress action toward a base
  \ElsIf{relocation option is active}
    \State Choose graph-progress action toward its goal
  \Else
    \State Maximize Eq.~\eqref{eq:ant} over admissible actions
  \EndIf
  \State Add the selected next position to $P$
\EndFor
\State Execute the joint action and advance the simulator
\end{algorithmic}\end{algorithm}

\begin{table}[t]
\caption{EARS parameters reported with the supplied results and matching the inspected implementation defaults. Values are verified against the selected EARS checkpoint archived with the latest study.}
\label{tab:earsparams}
\centering
\footnotesize
\setlength{\tabcolsep}{4pt}
\begin{tabular}{@{}lr@{}}
\toprule
Parameter & Value \\
\midrule
Eligible-check history capacity & 12 positions \\
First eligible simulator step & 7 \\
Stagnation unique-cell ratio & $\leq 0.46$ \\
Repeated-visit ratio & $\geq 0.54$ \\
Congestion threshold & $\geq 1.15$ \\
Relocation deadline offset & 9 steps (inclusive) \\
Cooldown deadline offset & $9+7$ steps \\
\bottomrule
\end{tabular}
\end{table}

\subsection{EARS Family and Scope of Comparison}
EARS-Safe, EARS-NP-Safe, H-MAPPO-EARS-Safe, and EARS-Info-v9 are members of our proposed method family. EARS-NP adds a decaying, diffusing repulsive pheromone field to discourage revisits and congestion. H-MAPPO-EARS uses a repository-trained MAPPO local option within a hierarchy of return, relocation, escape, and exploitation behaviors. It is a hybrid controller rather than a new MAPPO training algorithm. These earlier family members measure the development of the proposed approach and are not independent external competitors.

The three additional v9 variants are diagnostic controls of our information controller. EARS-Info-Local-v9 disables relocation while retaining the information-based local rule, graph return logic, monitor, and action ordering. EARS-Info-Periodic16 changes only the event-request schedule to a fixed period of 16 steps after initial eligibility, preserving routing and feasibility checks; accepted intervention counts are measured rather than assumed to match. EARS-Info-NoSensingReservations removes sensing-footprint overlap reservations while preserving collision-related position reservations. All four v9 variants use the same newly selected EARS-Safe checkpoint and the same routing configuration except for the specified intervention.

\subsection{Marginal Sensing Gain and Local Action Selection}
Let $\mathcal F(c)$ contain non-obstacle cells within Manhattan sensor radius $r$ of cell $c$, and let $\mathcal R$ denote already reserved sensing cells. The information field is
\begin{equation}
 G_{\mathcal R}(c)=\frac{\sum_{q\in\mathcal F(c)\setminus\mathcal R}\mathbb I[u(q)\geq0.99]}{1+2r(r+1)}.
 \label{eq:gain}
\end{equation}
This is a normalized count of unobserved cells, not an entropy estimate or access to hidden mission priorities. The fixed denominator is the full sensor footprint, including at boundaries where fewer cells can be sensed. The field is computed by summing shifted uncertainty arrays over the sensor diamond.

For an admissible action with destination $c$, v9 uses
\begin{equation}
\begin{aligned}
 A_i^{\rm info}(c)={}&2G_{\mathcal R}(c)+\frac{0.35}{1+v(c)}-0.10\min(v(c),5)\\
 &-0.15\kappa_i(c)-0.02\mathbb I[a_i=\textsc{Stay}],
\end{aligned}\label{eq:infoaction}
\end{equation}
where $\kappa_i$ sums inverse Manhattan distances to at most the three nearest other robots within four cells, using a minimum denominator of 0.5. Safe actions are evaluated directly and ties follow the action label. After each robot's decision, its intended next sensing footprint enters $\mathcal R$. Collision reservations remain a separate part of the monitor. Thus, v9 changes local information collection as well as relocation and should not be interpreted as an event-only modification of EARS-Safe.

\subsection{Event Requests and Feasible Information Routing}
V9 records positions once per simulator step. Once the 12-position history is full, a unique-position fraction at most 0.46 requests reallocation. A second request condition is exhausted local information: the maximum unreserved $G(c)$ over the current cell and graph neighbors is zero. Requests begin at step seven, require online status, and respect retry and cooldown intervals. Robots returning to base or already following a route do not issue new requests.

Let $d_i(c)$ be shortest-path distance from robot $i$ to $c$ on the traversable grid, and $d_B(c)$ the graph distance to a reachable base. A candidate satisfies
\begin{equation}
\begin{gathered}
2\leq d_i(c)\leq\min(32,T_{\max}-t),\\
0.55\,[d_i(c)+d_B(c)]+18\leq b_i(t).
\end{gathered}\label{eq:feasible}
\end{equation}
Its utility is
\begin{equation}
 V_i(c)=\frac{2G_{\mathcal R}(c)+0.35/(1+v(c))}{1+0.15d_i(c)}.
 \label{eq:utility}
\end{equation}
Candidates with zero gain or failed feasibility are discarded. The best candidate is accepted only if $V_i(c)$ exceeds the best current/neighbor numerator by at least 0.08. Otherwise, the robot keeps its local information policy. This compares discounted distant opportunity with immediately accessible information; the threshold is fixed rather than selected on test outcomes.

Robots requesting routes are considered in ascending battery and identifier order. Existing active goals reserve their sensing footprints before new goals are allocated. An accepted route receives a deadline based on its graph distance plus four slack steps, capped by the remaining horizon. Arrival, expiry, exhausted goal information, inactivity, or return necessity cancels it. Accepted and rejected requests receive a five-step retry backoff; arrival/cancellation also starts a five-step cooldown. During route following, graph progress has priority over information gain and visit-count tie breakers, preventing accumulated visits from reversing the path objective.

Return behavior activates when $b_i\leq0.55(d_B(x_i)+1)+18$. V9 uses a multi-source base-distance field, although the surrounding runtime monitor retains its native checks. The graph-based reserve improves route screening but is not a proof of recursive feasibility.

\begin{algorithm}[t]
\caption{EARS-Info-v9 decision at step $t$}\label{alg:info}
\begin{algorithmic}[1]\small
\State Verify cached topology; update gain field and position histories
\State Cancel finished, stale, exhausted, or return-preempted routes
\For{each active robot eligible to request relocation}
\State Test stagnation and local information exhaustion
\EndFor
\State Reserve sensing around active goals
\For{requesting robots in battery/identifier order}
\State Build/reuse graph distances and apply Eq.~\eqref{eq:feasible}
\State Accept best Eq.~\eqref{eq:utility} only above the local margin
\State Reserve accepted goal footprint; set retry/deadline
\EndFor
\For{robots in identifier order}
\State Form admissible actions using planned-position reservations
\State Select inactive stay, return, route progress, or Eq.~\eqref{eq:infoaction}
\State Reserve selected position and sensing footprint
\EndFor
\State Execute the joint action
\end{algorithmic}\end{algorithm}

\subsection{Caching and Computational Cost}
For $M=L^2$ cells, the traversable graph has $O(M)$ vertices and edges. A breadth-first distance field costs $O(M)$; a multi-source traversal supplies base distances. Topology is keyed by environment identity, grid size, obstacles, restricted cells, and bases. An ordered cache retains at most 96 goal-distance fields across steps and is invalidated on topology changes. Its worst-case storage is $O(96M)$. Information-field convolution costs $O(Mr^2)$ for fixed sensor radius. Each relocation request still scans grid candidates and may build a distance field. Caching reduces repeated graph work; it does not imply constant-time planning or a general asymptotic speedup. Reported wall-clock gains below are empirical on the recorded platform.

\begin{figure*}[t]\centering
\includegraphics[width=0.95\textwidth]{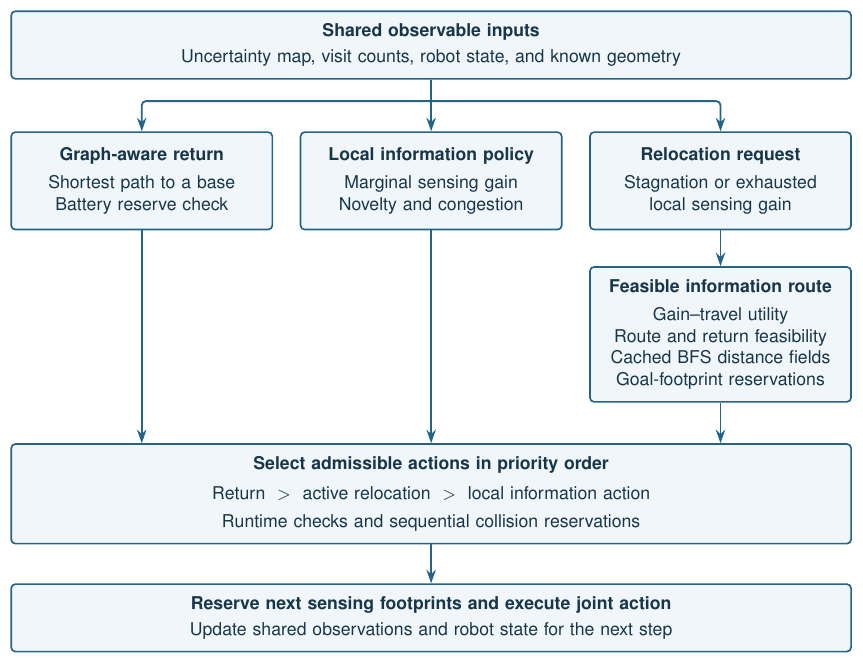}
\caption{The proposed information controller. Both local exploration and relocation use observable sensing gain. Graph return and runtime action filtering remain part of execution.}\label{fig:infoarchitecture}
\end{figure*}

\section{Experimental Methodology}
\subsection{Geography, Observation, and Map Provenance}
Training uses Vienna, London, and Berlin; validation uses Amsterdam and Prague. Evaluation uses San Francisco, Paris, Barcelona, Rome, Manhattan/New York, Chicago, Tokyo, and Melbourne. Training start zones are north-west, south-east, and center; validation uses west, east, north, and south; evaluation alternates north-east and south-west. The current training and checkpoint selection do not consume evaluation metrics. Because city identities were used in earlier studies, these splits do not erase development exposure to the geography.

Each snapshot queries OpenStreetMap around a geocoded center with radius parameter 1.4~km and maps returned features to a $40\times40$ grid \cite{OSMContributors}. Points or representative points are scaled by the returned feature bounds. Buildings mark obstacles; industrial/railway land use and water/wetland mark restrictions. Mission priorities are 1.00 for emergency/clinical facilities, 0.85 for schools/care/community facilities, 0.75 for transport, 0.65 for tourism, 0.55 for recreation, and 0.45 for shops/offices. Co-located targets retain the maximum weight; mission and base cells override blocked labels. Start preprocessing selects traversable bases near the assigned zone.

Geometry and mission provenance must both be OpenStreetMap-derived. Invalid legacy caches are refreshed; synthetic fallback, empty missions, invalid priorities, out-of-grid cells, and target/blocked overlap are rejected. All methods then use identical frozen snapshots. These checks establish provenance and internal consistency, not fidelity to building footprints or physical mobility. Table~\ref{tab:maps} reports actual grid complexity. Chicago is particularly sparse; it remains in the original eight-map descriptive panel, with exclusion sensitivity reported separately. No map was dropped after seeing method scores.
\begin{table}[t]
\centering\footnotesize
\caption{Frozen test-map abstraction. Reachability is from the north-east start; Tokyo reaches 100\% of targets from the south-west start.}\label{tab:maps}
\setlength{\tabcolsep}{3pt}
\resizebox{\columnwidth}{!}{\begin{tabular}{@{}lrrr@{}}\toprule
City & Targets & Obstacles & Reachable (\%) \\ \midrule
San Francisco & 879 & 241 & 99.20 \\
Paris & 236 & 2 & 100.00 \\
Barcelona & 476 & 30 & 100.00 \\
Rome & 531 & 139 & 99.81 \\
New York & 153 & 4 & 100.00 \\
Chicago & 1 & 0 & 100.00 \\
Tokyo & 509 & 118 & 99.61 \\
Melbourne & 313 & 46 & 100.00 \\
\bottomrule\end{tabular}}
\par\smallskip\parbox{\linewidth}{\scriptsize The grid has 1,600 cells. Counts reflect merged representative-point labels, not building footprints. Chicago has two restricted cells despite having no obstacle cells.}
\end{table}

\subsection{Training, Selection, and Method Ownership}
The learned GRPO, IPPO, MAPPO, and HAPPO implementations are trained on the development cities, followed by the existing teacher-assisted coordination upgrade. PRISM configurations and the EARS family are selected using training and validation performance. The initial learning command requests 108 episodes per learned policy, 12 validation episodes and two validation repeats; coordination uses two rounds; PRISM uses eight rounds; EARS shortlisting uses six training episodes and eight validation episodes with two repeats. Commands, selection histories, checkpoint hashes, and elapsed times are archived. These budgets describe the pipeline rather than a claim of equal optimizer steps across methods. One training replicate is available.

The comparison includes 26 preceding implementations and four v9 implementations. The earlier EARS methods, the proposed v9 controller, and its controls are our family of methods; their progression is shown explicitly. Other implementations provide comparison context. QMIX-Safe, MADDPG-Safe, and MAT-Safe are lightweight policy-style adaptations, not complete published QMIX, MADDPG, or MAT reproductions. Repository PPO-family policies use trained linear/residual models. The comparison therefore concerns the frozen implementations, not a general ranking of published MARL algorithms. RandomAgent and GreedyAgent are unfiltered reference controls. Heuristics have no learned weights; GRPO ablations share the corresponding trained checkpoint. Native filter and return implementations differ.

\subsection{Evaluation Grid and Paired Execution}
Both mission-stop and fixed-budget panels evaluate $N\in\{4,8,16,32\}$ robots and dropout $p_d\in\{0,0.03,0.15,0.30\}$, with 20 episodes per city/setting/method. Each panel contains $8\times16\times20\times30=76,800$ episodes. Mission-stop ends at complete discovery, all-agent termination, or step 160. Fixed-budget disables mission-completion stopping, but all-agent termination can still shorten a rollout. Consequently, it is a horizon-controlled rather than guaranteed equal-duration comparison.

The SWAP panel uses eight robots, 3\% dropout, three target seeds (2042, 3051, 4060), and 20 episodes per city/seed/method, giving 14,400 episodes. Total evaluation size is 168,000. The primary setting is mission-stop with eight robots and 3\% dropout, giving 160 episodes per method and 4,800 episodes across methods. Each episode starts robots with 100 battery units, uses movement cost 0.55, idle cost 0.10, Manhattan sensing radius one, sensed uncertainty factor 0.25, and pheromone decay 0.96.

All methods share frozen maps, starts, episode seeds, and communication draws indexed by simulator step and robot. Draw hashes agree within every one of the 5,600 matched scenarios. Method execution is serial with rotated method order. An observed-only environment interface hides unsensed mission labels; this is verified through action-parity fixtures. Shared map access nevertheless persists during disconnection, so dropout is not a realistic delayed decentralized map-exchange model.

\subsection{Primary Inference and Secondary Analyses}
The frozen primary contrast is EARS-Info-v9 minus EARS-Safe in operational score at the primary setting. For city $c$, average paired episode differences to obtain $\Delta_c$; report $\bar\Delta=\frac18\sum_c\Delta_c$. The exact two-sided sign-flip test enumerates all $2^8$ assignments of city signs and compares absolute means. Its interpretation depends on the symmetry/exchangeability assumption for city contrasts. It does not make eight reused cities a random representative sample of all environments.

A descriptive city-bootstrap interval resamples the eight city means 20,000 times with replacement (seed 91827). It measures between-city variation conditional on the frozen checkpoints, not training-seed variability. Equal-city aggregation is retained for all panel means. The 1,739 secondary comparisons cover metrics, comparators, and panels, and use one conservative Holm family. They are exploratory; the primary test is reported separately. At eight cities, the minimum two-sided sign-flip $p$ is 0.0078125, so the broad correction also limits power for component-level conclusions. The family is not narrowed retrospectively to obtain significance.

SWAP changes hidden mission subsets on fixed geometry. Seven cities have distinct changed views for all three seeds. Chicago's one-target view is unchanged and repeated; it remains in the descriptive all-city SWAP panel but is excluded from changed-view inference. We additionally examine a Chicago-exclusion primary effect and score-weight sensitivity using the same frozen episode data. These are post hoc robustness descriptions, not additional independent experiments.

\section{Results}
\subsection{Complete-Controller Improvement at the Primary Setting}
Table~\ref{tab:latestall} reports every implementation, with the authors' EARS family and its controls marked explicitly. EARS-Info-v9 ranks first at $J=0.814958$, compared with $0.772998$ for EARS-Safe: an absolute improvement of $0.041960$, or 5.43\% relative. The exact city sign-flip test gives $p=0.0078125$ and the descriptive 95\% city-bootstrap interval is $[0.015908,0.069364]$. All eight city contrasts are positive. This supports the complete v9 implementation over the preceding EARS-Safe controller under the evaluated conditions.

Weighted discovery rises from 0.888788 to 0.984619, an improvement of 9.583 percentage points. Coverage changes from 0.538550 to 0.542286, a smaller 0.374-point increase. Mean energy falls from 574.3625 to 559.3222 units (2.62\%), and complete-target episodes increase from 31.25\% to 35.625\%. Mean capped completion steps decrease from 138.6875 to 134.0875. These are mission-stop rollout outcomes, not matched-duration energy savings. Runtime is reported separately from $J$ and cannot account mechanically for the score increase.
\begin{table*}[t]
\centering\footnotesize
\caption{All 30 implementations at eight robots, 3\% dropout, mission stopping. Equal-city means over 160 episodes per method.}\label{tab:latestall}
\setlength{\tabcolsep}{4pt}
\begin{tabular}{@{}lrrrrrr@{}}\toprule
Method & $J$ & Discovery (\%) & Coverage (\%) & Energy & Time (s) & Complete (\%) \\ \midrule
\textbf{EARS-Info-v9$^{\dagger}$} & \textbf{0.8150} & \textbf{98.46} & \textbf{54.23} & \textbf{559.32} & \textbf{0.420} & \textbf{35.62} \\
EARS-Info-NoSensingReservations$^{\dagger}$ & 0.8117 & 97.71 & 54.54 & 566.77 & 0.418 & 31.87 \\
EARS-Info-Periodic16$^{\dagger}$ & 0.8021 & 94.38 & 55.08 & 565.89 & 0.436 & 33.12 \\
H-MAPPO-EARS-Safe$^{\dagger}$ & 0.7843 & 91.24 & 56.44 & 597.57 & 5.983 & 25.00 \\
PRISM-Plus-Safe & 0.7789 & 92.37 & 53.98 & 605.73 & 4.581 & 25.00 \\
EARS-NP-Safe$^{\dagger}$ & 0.7785 & 89.87 & 57.03 & 607.59 & 3.295 & 18.75 \\
AntSwarmSafe & 0.7760 & 90.14 & 54.68 & 588.26 & 2.125 & 12.50 \\
EARS-Safe$^{\dagger}$ & 0.7730 & 88.88 & 53.85 & 574.36 & 3.089 & 31.25 \\
PRISM-X-Safe & 0.7546 & 87.83 & 51.78 & 604.24 & 4.513 & 18.75 \\
PRISM-Ant-Safe & 0.7533 & 86.73 & 53.55 & 620.18 & 5.161 & 18.75 \\
PRISM-Safe & 0.7479 & 84.67 & 53.65 & 617.70 & 4.685 & 18.75 \\
PRISM-Star-Safe & 0.7479 & 84.67 & 53.65 & 617.70 & 4.700 & 18.75 \\
MAPPO-Safe & 0.7189 & 82.91 & 44.93 & 583.35 & 3.905 & 18.75 \\
IPPO-Safe & 0.7048 & 77.75 & 46.04 & 586.35 & 3.888 & 12.50 \\
HAPPO-Safe & 0.6512 & 60.59 & 43.07 & 527.94 & 4.064 & 6.25 \\
BeeSwarmSafe & 0.6274 & 52.98 & 27.00 & 330.67 & 2.201 & 12.50 \\
EARS-Info-Local-v9$^{\dagger}$ & 0.6182 & 53.31 & 46.42 & 592.62 & 0.192 & 12.50 \\
UA-HBAS-Safe & 0.6045 & 42.37 & 27.66 & 324.04 & 2.438 & 0.00 \\
GRPO-Safe-Ablation-NoPropagation & 0.5936 & 48.56 & 35.26 & 499.77 & 4.378 & 12.50 \\
PSOSwarmSafe & 0.5914 & 44.27 & 26.69 & 361.71 & 0.265 & 12.50 \\
GRPO-Safe-Ablation-NoMemory & 0.5815 & 44.20 & 35.60 & 494.19 & 4.329 & 12.50 \\
SafeSwarmAgent & 0.5811 & 43.75 & 31.50 & 420.12 & 2.072 & 12.50 \\
GRPO-Safe & 0.5493 & 39.27 & 33.26 & 508.22 & 4.711 & 5.62 \\
SafetyFilteredGreedy & 0.5263 & 32.86 & 25.60 & 404.22 & 2.154 & 6.25 \\
GRPO-Safe-Ablation-NoLearnedBehavior & 0.5168 & 29.76 & 30.92 & 470.42 & 4.656 & 6.25 \\
MADDPG-Safe & 0.4689 & 31.14 & 27.98 & 584.69 & 2.747 & 1.25 \\
GreedyAgent & 0.4492 & 32.04 & 25.64 & 474.47 & 2.035 & 6.25 \\
MAT-Safe & 0.4391 & 24.92 & 25.02 & 584.77 & 2.855 & 0.62 \\
QMIX-Safe & 0.4186 & 20.85 & 22.56 & 575.53 & 2.799 & 0.00 \\
RandomAgent & 0.3536 & 9.41 & 14.58 & 559.86 & 0.034 & 0.00 \\
\bottomrule\end{tabular}
\par\smallskip\parbox{\linewidth}{\scriptsize $\dagger$: authors\textquotesingle{} EARS family or matched v9 control. Other rows include repository heuristics, hybrids, trained policies, and adaptations; they are not all external reproductions. All filtered methods have zero mean actual incidents at this setting. RandomAgent and GreedyAgent have 28.4813 and 388.5625, respectively. Complete is the fraction of episodes discovering all targets; discovery is weighted target discovery.}
\end{table*}

\subsection{Performance Across Evaluation Panels}
EARS-Info-v9 has the highest equal-city mean score in the full mission-stop, fixed-budget, all-city SWAP, and changed-view SWAP panels (Table~\ref{tab:panels}). The overall mission-stop mean averages all 16 team/dropout settings and differs from the primary-setting mean. Comparisons between stopping rules remain separate; pooling their episodes would obscure the operational meaning of energy and completion.

\begin{figure*}[t]\centering
\includegraphics[width=.98\textwidth]{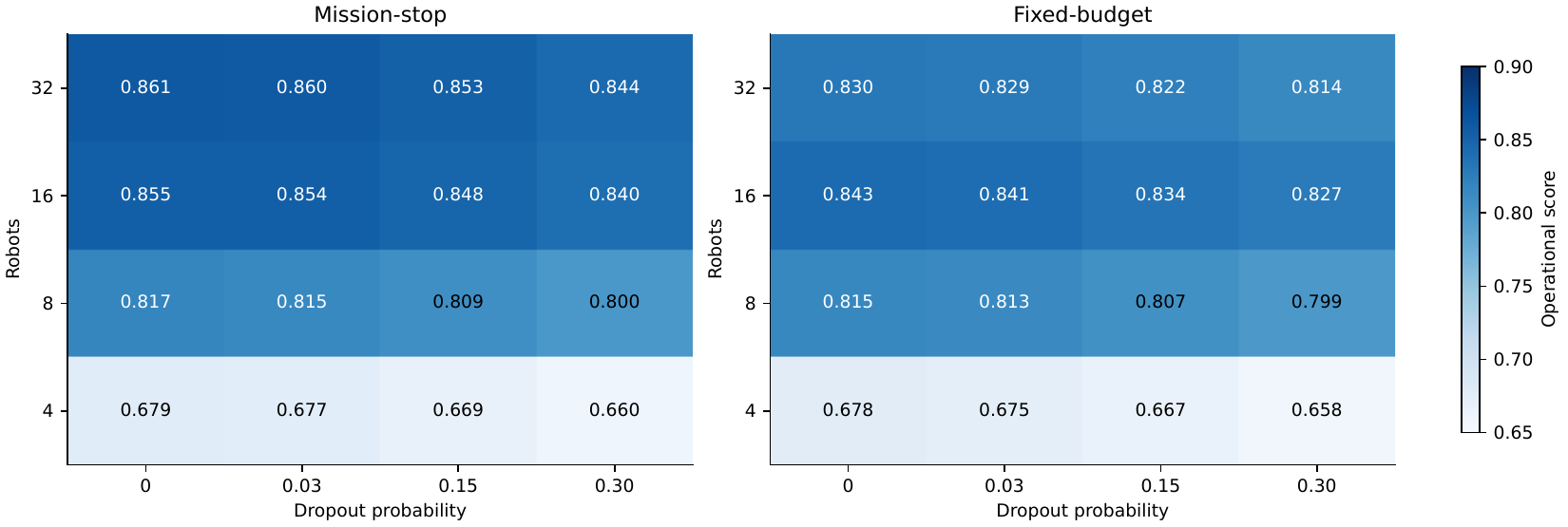}
\caption{EARS-Info-v9 operational score across all tested team/dropout settings. Each cell is the equal-city mean over 20 episodes per city. Mission-stop and fixed-budget results are kept separate.}\label{fig:settings}
\end{figure*}

Figure~\ref{fig:settings} displays the measured score across team sizes and dropout settings. These are evaluations of one fixed configuration, without retuning for individual test cities or settings. Higher dropout does not amount to full communication isolation because shared observations remain accessible. The results characterize robustness to this simulator's dropout process.
\begin{table}[t]
\centering\footnotesize
\caption{Selected family and reference scores across panels. All 30 methods are retained in the accompanying CSV.}\label{tab:panels}
\setlength{\tabcolsep}{3pt}
\resizebox{\columnwidth}{!}{\begin{tabular}{@{}lrrrr@{}}\toprule
Method & Mission & Fixed & SWAP & Changed \\ \midrule
EARS-Info-v9 & 0.7963 & 0.7846 & 0.8154 & 0.8161 \\
EARS-Safe & 0.7711 & 0.7607 & 0.7818 & 0.7777 \\
AntSwarmSafe & 0.7716 & 0.7633 & 0.7738 & 0.7685 \\
H-MAPPO-EARS & 0.7757 & 0.7673 & 0.7735 & 0.7670 \\
Periodic16 & 0.7816 & 0.7723 & 0.8023 & 0.8010 \\
No sensing reservations & 0.7930 & 0.7821 & 0.8126 & 0.8129 \\
\bottomrule\end{tabular}}
\par\smallskip\parbox{\linewidth}{\scriptsize Mission/fixed average 16 settings; SWAP uses eight robots and 3\% dropout. Changed excludes the unchanged Chicago target views.}
\end{table}

\subsection{Geographic Heterogeneity and Sparse-Map Sensitivity}
\begin{figure}[t]\centering
\includegraphics[width=\columnwidth]{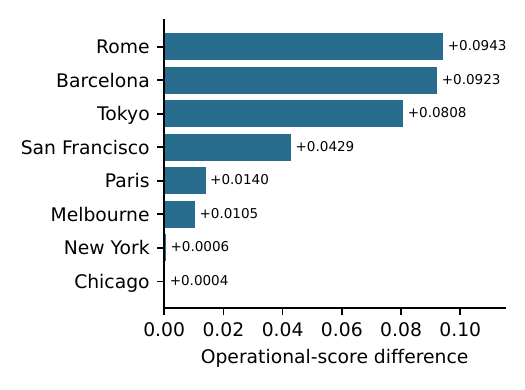}
\caption{Primary-setting city mean score differences, EARS-Info-v9 minus EARS-Safe. All eight are positive; magnitudes vary considerably.}\label{fig:latestcity}
\end{figure}

The primary city contrasts range from $+0.000373$ in Chicago and $+0.000649$ in New York to $+0.094265$ in Rome and $+0.092255$ in Barcelona (Fig.~\ref{fig:latestcity}). Tokyo improves by $+0.080779$, San Francisco by $+0.042886$, Paris by $+0.014024$, and Melbourne by $+0.010451$. Thus, the improvement is directionally consistent but varies substantially in magnitude.

Excluding Chicago gives a mean score improvement of $+0.047901$ across seven cities. This post hoc sensitivity shows that the primary gain is not created by the sparse Chicago instance; it does not justify deleting the city from the original analysis. The small Manhattan effect and near-empty obstacle grids in some snapshots also caution against equating a city label with equivalent geometric difficulty. Comparisons are paired within the exact abstraction rather than across supposedly standardized physical urban environments.

\subsection{Matched Controls and Limits of Mechanism Evidence}
Disabling relocation gives $J=0.618154$, versus 0.814958 for the complete v9 controller; the city contrast is $+0.196804$ with nominal $p=0.015625$. This sizeable difference is compatible with a useful role for relocation in the information-policy architecture. However, its adjusted secondary $p$ is 1.0, so it is not an adjusted confirmatory claim.

Periodic relocation yields $J=0.802061$ and the no-sensing-reservation variant yields $0.811736$. Full v9 exceeds them by $0.012897$ and $0.003222$, with nominal city $p=0.125$ and $0.25$, respectively (Table~\ref{tab:controls}). Neither difference establishes that event timing or sensing reservations independently improve generalization. Periodic and event-driven methods do not have identical accepted intervention budgets. No secondary comparison survives the specified Holm family. The complete-controller result and component-level uncertainty can coexist: the primary improvement is a joint effect of local scoring, routing, return behavior, caching, and coordination.
\begin{table}[t]
\centering\footnotesize
\caption{Primary-setting EARS-Info-v9 score contrasts. The EARS-Safe contrast is primary; all others are secondary.}\label{tab:controls}
\setlength{\tabcolsep}{3pt}
\resizebox{\columnwidth}{!}{\begin{tabular}{@{}lrrr@{}}\toprule
Comparator & $\Delta J$ & Nominal $p$ & Holm $p$ \\ \midrule
EARS-Safe & +0.0420 & 0.0078 & -- \\
AntSwarmSafe & +0.0390 & 0.0156 & 1.00 \\
H-MAPPO-EARS & +0.0307 & 0.0234 & 1.00 \\
Periodic16 & +0.0129 & 0.1250 & 1.00 \\
No reservations & +0.0032 & 0.2500 & 1.00 \\
Local only & +0.1968 & 0.0156 & 1.00 \\
\bottomrule\end{tabular}}
\end{table}

\subsection{Runtime and Simulated Incidents}
At the primary setting, EARS-Info-v9 requires 0.419947~s per episode on average, compared with 3.089217~s for EARS-Safe, a measured ratio of 7.36. Policy time changes from 3.064597 to 0.399838~s. The comparison is conditional on the recorded hardware, serial execution, and stopping rule. Caching provides a plausible implementation-level explanation, but a matched cache-disabled experiment would be needed to attribute this ratio to caching alone.

EARS-Info-v9 records zero collisions, restricted-zone entries, battery failures, and combined actual incidents across its 5,600 evaluations. The filtered competitors also record zero combined incidents, whereas unfiltered controls do not. This establishes successful simulated execution in the evaluated episodes; it does not demonstrate superior safety over the filtered methods or guarantee physical safety.

\subsection{Score Dependence}
The composite score is fixed before evaluation and excludes runtime. Raw discovery, coverage, energy, completion, and incident metrics accompany it. The principal improvement is not merely a runtime-weight artifact: runtime has zero weight, weighted discovery increases markedly, energy decreases, and coverage increases slightly. Nevertheless, the ordering depends on chosen operational preferences. We drew 2,000 independent weight vectors by multiplying each original score coefficient by a uniform factor in $[0.75,1.25]$ and renormalizing the six coefficients (seed 91827). EARS-Info-v9 remains above EARS-Safe and ranks first among all 30 implementations for every tested vector. Its mean-score margin over EARS-Safe ranges from $0.03333$ to $0.05073$. These are fixed-data preference sensitivities, not independent samples, additional simulation episodes, or new significance tests.

\section{Discussion}
\subsection{What Changes Relative to Earlier EARS}
The central finding is an improvement of the complete information-aware controller over the preceding EARS-Safe implementation, with positive score differences in each of the eight evaluated abstractions. Earlier EARS uses observable evidence, novelty, and event-triggered global relocation. V9 directly values newly sensed cells, screens routes using graph travel and return distances, coordinates marginal sensing, and reuses graph structure. The large discovery improvement with a small coverage change suggests that where sensing occurs matters beyond the count of visited cells. This interpretation is consistent with the design but is not an independently identified causal decomposition.

The comparison among our EARS family members is deliberately distinct from the comparison with other implementations. H-MAPPO-EARS and EARS-NP represent alternative extensions, not external claims against which novelty is manufactured. The matched v9 controls ask narrower design questions. Other swarm, PRISM, and learned-policy implementations establish broader benchmark context, while adapted QMIX/MADDPG/MAT labels must not be read as full external algorithm reproductions.

\subsection{Practical Interpretation}
Within this static, known-geometry benchmark, EARS-Info-v9 offers a favorable measured combination of discovery, energy, and runtime. It is appropriate when the decision system has access to shared uncertainty/visit fields and a traversable graph. The reported benefit does not directly transfer to unknown geometry, delayed map exchange, uncertain localization, continuous dynamics, or real battery discharge. Those settings require an observation/communication model and an execution layer beyond this study.

\section{Limitations}
The study is exploratory because evaluation cities were previously used during development. Coarse rasterization produces sparse maps, including Chicago's single-target instance. One training replicate and unequal model capacity and training resources limit generalization and external MARL comparisons. Joint design changes prevent isolated causal attribution, and no secondary test survives multiplicity correction. Shared state persists during dropout, execution safeguards differ, and early termination limits energy comparisons. Runtime is platform-specific; zero simulated incidents do not guarantee physical safety.

\section{Reproducibility and Artifact Availability}
Code is available at \url{https://github.com/vtavakkoli/SafeSwarm-CityTwin}. The frozen \nolinkurl{eight-city-publication-v3} study contains 168,000 episode records covering 30 methods, maps, checkpoints, source snapshots, and training/runtime manifests. All 227 checked hashes matched; independent recalculation reproduced the scores, panel means, and primary $p$-value.

The runtime manifest records software versions, hardware, and single-thread settings; no immutable image digest was supplied. The CSV supports the reported analyses, but the omitted SQLite journal and full diagnostics are required for the original journal-based reporting workflow. Analysis scripts and an artifact inventory accompany the manuscript materials. OpenStreetMap data retain their separate attribution and licensing requirements \cite{OSMContributors}.

\section{Conclusion}
EARS-Info extends our EARS method family with marginal sensing gain, graph-feasible event-driven relocation, sensing reservations, and reusable path fields. In 168,000 matched evaluations of 30 implementations across eight OSM-derived city maps, EARS-Info-v9 achieves the highest mean operational score in every reported panel. At the primary setting, its score improves by 5.43\% over EARS-Safe, weighted discovery increases by 9.58 percentage points, and mean rollout energy decreases by 2.62\%. All eight city score contrasts are positive, with exact $p=0.0078125$. These findings support the complete-controller improvement in the tested benchmark. Component-level superiority, independent geographic generalization, and physical safety remain outside the established evidence. The result is an interpretable and auditable improvement with explicit implementation, statistical, and environmental scope.

\begingroup\small
\bibliographystyle{IEEEtran}
\bibliography{references}
\endgroup

\end{document}